\documentclass[10pt,journal,cspaper,compsoc]{IEEEtran}
\ifCLASSOPTIONcompsoc
\else
\fi
\ifCLASSINFOpdf
\else
\fi
\usepackage[cmex10]{amsmath}
\usepackage{algorithmic,epsfig,subfigure}
\begin{document}
\title{Kernel Reconstruction ICA for Sparse Representation}

\author{Yanhui~Xiao,
 Zhenfeng~Zhu,
 and~Yao~Zhao% <-this % stops a space
\thanks{Y. Xiao and Z. Zhu are with the Institute of Information Science, Beijing Jiaotong University, and with Beijing Key Laboratory of Advanced Information Science and Network Technology, Beijing 100044, China (e-mail: xiaoyanhui@gmail.com, zhfzhu@bjtu.edu.cn).}
\thanks{Y. Zhao is with the Institute of Information Science, Beijing Jiaotong University,
and with State Key Laboratory of Rail Traffic Control and Safety,
Beijing 100044, China (e-mail: yzhao@bjtu.edu.cn).}}


\markboth{Journal of \LaTeX\ Class Files,~Vol.~6, No.~1, January~2007}%
{Shell \MakeLowercase{\textit{et al.}}: Bare Demo of IEEEtran.cls for Computer Society Journals}

%\IEEEpubid{\makebox[\columnwidth]{\hfill 0000--0000/00/\$00.00~\copyright~2007 IEEE}%

% PRIOR to the title within the \IEEEcompsoctitleabstractindextext IEEEtran
\IEEEcompsoctitleabstractindextext{%
\begin{abstract}
%\boldmath
Independent Component Analysis (ICA) is an effective unsupervised tool to learn statistically independent representation.  However, ICA is not only sensitive to whitening but also difficult to learn an over-complete basis. Consequently, ICA with soft Reconstruction cost(RICA) was presented to learn sparse representations with over-complete basis even on unwhitened data. Whereas RICA is infeasible to represent the data with nonlinear structure due to its intrinsic linearity. In addition, RICA is essentially an unsupervised method and can not utilize the class information. In this paper, we propose a kernel ICA model with reconstruction constraint (kRICA) to capture the nonlinear features. To bring in the class information, we further extend the unsupervised kRICA to a supervised one by introducing a discrimination constraint, namely d-kRICA. This constraint leads to learn a structured basis consisted of basis vectors from different basis subsets corresponding to different class labels. Then each subset will sparsely represent well for its own class but not for the others. Furthermore, data samples belonging to the same class will have similar representations, and thereby the learned sparse representations can take more discriminative power. Experimental results validate the effectiveness of kRICA and d-kRICA for image classification.
\end{abstract}
% not want either math or citations to appear in the abstract.

% Note that keywords are not normally used for peer review papers.
}

\maketitle

% is not selected <OR> if conference mode is selected - because compsoc
% conference papers position the abstract like regular (non-compsoc)
\IEEEdisplaynotcompsoctitleabstractindextext

% \fi
\IEEEpeerreviewmaketitle

\section{Introduction}

Sparsity is an attribute characterizing a mass of natural and manmade signals~\cite{sparsity}, and has played a vital role in the success of many machine learning algorithms and techniques such as compressed sensing~\cite{donoho2006compressed}, matrix factorization~\cite{NMF}, sparse coding~\cite{sparsecoding}, dictionary learning~\cite{KSVD,DKSVD}, sparse auto-encoders~\cite{autoencoders}, Restricted Boltzmann Machines (RBMs)~\cite{RBMs} and Independent Component Analysis (ICA)~\cite{ICA2001}.

Among these, ICA transforms an observed multidimensional random vector into sparse components which are statistically as independent from each other as possible. Specifically, to estimate the independent components, a general principle is the maximization of non-gaussianity~\cite{ICA2001}. This is based on the central limit theorem that sum of independent random variables is closer to gaussian than any of the original random variables, i.e., non-gaussian is independent. Meanwhile, sparsity is one form of non-gaussianity~\cite{NaturalImageStatistics}, which is dominant in natural images. Then maximization of sparseness in natural images is basically equivalent to maximization of non-gaussianity. Thus, ICA has been successfully applied to learn sparse representation for classification tasks by maximizing sparsity~\cite{TICA}. However, there are two main drawbacks to standard ICA.

1) ICA is sensitive to whitening, which is an important preprocessing step in ICA to extract efficient features. In addition, standard ICA is difficult to exactly whiten high dimensional data. For example, an input image of size 100$\times$100 pixels could be exactly whitened by principal component analysis(PCA), while it has to solve the eigen-decomposition of the 10,000 $\times$ 10,000 covariance matrix.

2) ICA is hard to learn the over-complete basis (that is the number of basis vectors is greater than dimensionality of input data).  Whereas Coates et al.~\cite{coates2010analysis} have shown that several approaches with over-complete basis, e.g., sparse autoencoders~\cite{autoencoders}, K-means~\cite{coates2010analysis} and RBMs~\cite{RBMs}, obtain an improvement for the performance of classification. This puts ICA at a disadvantage compared to these methods.

Both drawbacks are mainly due to the hard orthonormality constraint in standard ICA. Mathematically, that is $WW^T = I$, which is utilized to prevent degenerate solution for the basis matrix $W$ where each basis vector is a row of $W$. While this orthonormalization cannot be satisfied when $W$ is over-complete. Specifically, the optimization problem of standard ICA is generally solved by using gradient descent methods, where $W$ is orthonormalized at each iteration by symmetric orthonormalization, i.e., $W \leftarrow (WW^T )^{-1/2}W$, which doesn't work for over-complete learning. In addition, although alternative orthonormalization methods could be employed to learn over-complete basis, they not only are expensive to compute but also may arise from the cumulation of errors.

To address the above issues, Q.V. Le et al.~\cite{Rica} replaced the orthonormality constraint with a robust soft reconstruction cost for ICA (RICA). Thus, RICA can learn sparse representation with highly over-complete basis even on unwhitened data. However, this model is so far also a linear technique which is infeasible to discover nonlinear relationships among input data.
Additionally, as an unsupervised method, RICA may not be sufficient for classification tasks, which failed to consider the association between the training sample and its class.

Recall that, to explore the nonlinear features, kernel trick~\cite{kernel} can be used to nonlinearly project the input data into a high dimensional feature space. Therefore, we develop a kernel extension of RICA (kRICA) to represent the data with nonlinear structure. In addition, to bring in label information, we further extend the unsupervised kRICA to a supervised one by introducing a discrimination constraint, namely d-kRICA. Particularly, this constraint maximizes the homogeneous representation cost and minimizes the inhomogeneous representation cost jointly, which leads to learn a structured basis consisted of basis vectors from different basis subsets corresponding to the class labels. Then each subset will sparsely represent well for its own class but not for the others. Furthermore, data samples belonging to the same class will have similar representations, and thereby the obtained sparse representation can take more discriminative power.

It is important to note that this work is fundamentally based on our previous work DRICA~\cite{dRICA}. In comparison to DRICA, we further improve our work as follows:

1) By taking advantage of the kernel trick, we replace the linear projection with nonlinear one to capture the nonlinear features. Experimental results show that our kernel extension usually further improves the image classification accuracy.

2) The discriminative capability of basis is further enhanced  by maximizing the homogeneous representation cost besides minimizing the inhomogeneous representation cost simultaneously. Thus, we can obtain a set of more discriminative basis vectors that are forced to sparsely represent better for their own classes but poorer for the others. Experiments show that this basis can further boost the performance for image classification.

%3) Motivated by the success of $L_2$ pooling~\cite{TCNN,pooling} for learning invariant features, we utilize $L_2$ pooling instead of the simple $L_1$ sparsity penalty, which encourages pooling together groups of related features to achieve complex invariances such as scale and rotational invariance.

3) In the experiments, we conduct comprehensive analysis for our proposed method, e.g., the effects of different parameters and kernels for image classification, experiment settings, and the similarity comparative analysis.

The rest of the paper is organized as follows.  In Section 2, we revisit related works on sparse coding and RICA, and describe the connection between them. Then we give a brief review of reconstruction ICA in Section 3. Section 4 introduces the details of our proposed kRICA, including its optimization problem and implementation. By incorporating the discrimination constraint, kRICA is further extended to supervised learning in Section 5. Section 6 presents extensive experimental results on image classification. Finally, we conclude our work in Section 7.

\section{Related Work}
In this section, we will review some related work in the following aspects: (1) Sparse coding and its applications; (2) Connection between RICA and sparse coding; (3) The other kernel sparse representation algorithms.

%Sparse coding is a method for seeking a representation of data in which each of the components of the representation is only rarely significantly active.

Sparse coding is an unsupervised method for reconstructing a given signal by selecting a relatively small subset of basis vectors from an over-complete basis set, and meanwhile making the reconstruction error as small as possible.
Because of its plausive statistical theory~\cite{donoho2006most}, sparse coding has attracted more and more attention from scientists in computer vision field. Meanwhile, it has been successfully used for more and more computer vision applications, e.g., image classification~\cite{scspm,ILCC,lcksvd}, face recognition~\cite{SRC}, image restoration~\cite{imageRestoration} etc.
This success is largely due to two factors:

1) The sparsity characteristic ubiquitously exists in many computer vision applications. For example, for image classification, the image components can be sparsely reconstructed by utilizing similar components of other images from same class~\cite{scspm}. Another example is face recognition. The face image to be tested can be accurately reconstructed by a few training images from the same category~\cite{SRC}. As a consequence, sparsity is the foundation for these applications based on sparse coding.

2) Images are often corrupted by noise, which may arise due to sensor imperfection, poor illumination or communication errors. While sparse coding can effectively select the related basis vectors to reconstruct the clean image, and meanwhile can deal with noise by allowing the reconstruction error and promoting sparsity. Therefore, sparse coding has been successfully applied to image denoising~\cite{eladSparse}, image restoration~\cite{imageRestoration} etc.

Similar to sparse coding, ICA with a reconstruction cost (RICA)~\cite{Rica} also can learn highly over-complete sparse representation. In addition, in~\cite{Rica}, it has been shown that RICA is mathematically equivalent to sparse coding if using explicit encoding and ignoring the norm ball constraint.

The above-mentioned studies only seek the sparse representations of the input data in the original data space, which are incompetent to represent the data with nonlinear structure. To solve this problem, Yang et al.~\cite{kica} developed a two-phase kernel ICA algorithm: whitened kernel principal component analysis (KPCA) plus ICA. Different from~\cite{kica}, another solution~\cite{bachKICA} was proposed to use contrast function based on canonical correlations in a reproducing kernel Hilbert space. However, both of these methods couldn't learn the over-complete sparse representation of nonlinear features due to the orthonormality constraint. Therefore, to find such representation, Gao et al.~\cite{KSR_eccv10,KSR} presented a kernel sparse coding method (KSR) in a high dimensional feature space. But this work failed to utilize the class information as an unsupervised approach. Additionally, in Section 4.4, we will show that our proposed kernel extension of RICA (kRICA) is equivalent to KSR under certain conditions.

\section{Reconstruction ICA}
Since sparsity is one form of non-gaussianity, maximization of sparsity for ICA is equivalent to maximization of independence\cite{NaturalImageStatistics}. Given the unlabeled data set $X=\{ {x_{i}}\} _{i = 1}^m$ where $x_{i} \in R^n$, the  optimization problem of standard ICA~\cite{ICA2001} is generally defined as
\begin{equation}
\begin{array}{l}
 \mathop {\min}\limits_W \sum\limits_{i = 1}^m {\sum\limits_{j = 1}^K {g({w_j}{x_{i}}})}  \\
 s.t.\begin{array}{*{20}{c}}
   {} & {W{W^T}}  \\
\end{array} = I,\\
\end{array}
\end{equation}
where $g(\cdot)$ is a nonlinear convex function, $W=[w_1,w_2,\ldots,w_K]^T \in R^{K \times n}$ is the basis matrix, $K$ is the number of basis vectors and $w_j$ is $j$-th row basis vector in $W$, and $I$ is the identity matrix. Additionally, the orthonormality constraint $WW^T = I$ is traditionally utilized to prevent the basis vectors in $W$ from becoming degenerate. Meanwhile, a good general purpose smooth $L_1$ penalty is: $g(\cdot) = \log(\cosh(\cdot))$~\cite{NaturalImageStatistics}.

However, as above pointed out, the orthonoramlity constraint makes standard ICA difficult to learn the over-complete basis. In addition, ICA is sensitive to whitening. These drawbacks restrict ICA to scale high dimensional data. Consequently, RICA~\cite{Rica} used a soft reconstruction cost to replace the orthonormality constraint in ICA. Applying this replacement to Equation (\ref{rica}), RICA can be formulated as the following unconstrained problem
\begin{equation}
\label{rica}
\mathop {\min}\limits_W \frac{1}{m}\sum \limits_{i = 1}^m [{||{W^T}W{x_i} - {x_i}||_2^2 } + {\lambda{g({W}{x_{i}}})}],
\end{equation}
where parameter $\lambda$ is a tradeoff between reconstruction and sparsity. Swapping the orthonormality constraint with a reconstruction penalty, the RICA could learn sparse representations even on the data without whitening when $W$ is over-complete.

Furthermore, since the $L_1$ penalty is not sufficient to learn invariant features~\cite{NaturalImageStatistics}, RICA~\cite{Rica,le2012building} replaced it by a $L_2$ pooling penalty which encourages pooling features to group similar features together to achieve complex invariances such as scale and rotational invariance. Besides, the $L_2$ pooling can also promote sparsity for feature learning.
Particularly, $L_2$ pooling~\cite{l2pooling,pooling} is a two-layered network with square nonlinearity in the first layer, and square-root nonlinearity in the second layer:
\begin{align}
\label{equ:g}
{g({{W}}{x_{i}})} =\sum\limits_{j = 1}^K \sqrt {\varepsilon  + H_j{{({{W}}{x_{i}})}^2}},
\end{align}
where $H_j$ is the row of spatial pooling matrix $H \in R^{K \times K}$ fixed to uniform weights and $\varepsilon$ is a small constant to prevent division by zero.

Nevertheless, RICA is infeasible to represent the data with nonlinear structure due to its intrinsic linearity. In addition, this model just simply learned the over-complete basis set with reconstruction cost while failed to consider the association between the training sample and its class, which may be insufficient for classification tasks.
To address these problems, on one hand, we focus on developing a kernel extension of RICA to find the sparse representation of nonlinear features. On the other hand, we aim to learn a more discriminative basis by bringing in class information than unsupervised RICA, which will facilitate the better performance of sparse representation in classification tasks.

\section{Kernel Extension for RICA}
Motivated by the success that kernel trick can capture the nonlinear structure in data~\cite{kernel}, we propose a kernel version of RICA, called kRICA, to learn the sparse representation of nonlinear features.
\subsection{Model Formulation}
 Suppose that there is a kernel function $\kappa(\cdot,\cdot)$ induced by a high dimensional feature mapping $\phi: R^n \rightarrow R^\mathcal{N}$, where $n \ll \mathcal{N}$. Given two data points $x_i$ and $x_j$, $\kappa(x_{i},x_{j})={\phi(x_{i})}^T {\phi(x_{j})}$ represents a nonlinear similarity between them. Then the function maps the data and basis from the original data space to the feature space as follows.
\begin{equation}
\label{mapping}
\begin{gathered}
x\xrightarrow{\phi }{\phi(x)}\\
W=[w_1,\ldots,w_K]^T \xrightarrow{\phi } \mathcal{W} =[\phi({w_1}),\ldots,\phi({w_K})]^T
\end{gathered}
\end{equation}
Furthermore, by substituting the mapped data and basis into Equation (\ref{rica}), we can get the following objective function of kRICA.
\begin{equation}
\label{krica}
\mathop {\min}\limits_\mathcal{W} \frac{1}{m}\sum \limits_{i = 1}^m [{||{{\mathcal{W}}^T}{\mathcal{W}}{\phi(x_{i}}) - {\phi(x_{i})}||_2^2 } + \lambda{g({\mathcal{W}}\phi{(x_{i}}))}]
\end{equation}
Due to its excellent performance in many computer vision applications~\cite{kernel, KSR_eccv10}, Gaussian kernel, i.e., $\kappa(x_{i},x_{j})=\exp(-\gamma||x_{i}-x_{j}||_2^2)$ is used in this study. Thus, the norm ball constraints on basis in RICA can be removed owing to ${\phi(w_{i})}^T {\phi(w_{i})}=\kappa(w_{i},w_{i})=1$.

In addition, we perform kernel principal component analysis (KPCA) in the feature space for data whitening similar to~\cite{kica}, which makes the problem of ICA estimation simpler and better conditioned~\cite{NaturalImageStatistics}.
When data is whitened, there exists a close relationship between kernel ICA~\cite{kica} and kRICA. Regarding this relationship, we have the following Lemma:

\noindent \textbf{Lemma 4.1} \emph{When the input data set $X=\{ {x_{i}}\} _{i = 1}^m$ is whitened in the feature space,  the reconstruction cost $\frac{1}{m}\sum \limits_{i = 1}^m {||{{\mathcal{W}}^T}{\mathcal{W}}{\phi(x_{i}}) - {\phi(x_{i})}||_2^2 }$ is equivalent to the orthonormality cost ${||{{\mathcal{W}}^T}{\mathcal{W}} - I||_\mathcal{F}^2 }$.}

\noindent Where ${||\cdot||_\mathcal{F} }$ is the Frobenius norm. Lemma 4.1 shows that kernel ICA's hard orthonormality constraint and kRICA's reconstruction cost are equivalent when data is whitened. While kRICA can learn the over-complete sparse representation of nonlinear features and kernel ICA fails to work due to the orthonormality constraint. Please see the \textbf{Appendix A} for a detailed proof.

\subsection{Implementation}
The Equation (\ref{krica}) is an unconstrained convex optimization problem. To solve this problem, we rewrite the objective as follows
\begin{equation}
\label{equ:rewrite}
\begin{aligned}
&f(W)=\frac{1}{m}\sum\limits_{i = 1}^m [||{\mathcal{W}^T}\mathcal{W}\phi ({x_i}) - \phi ({x_i})||_2^2 + \lambda{g({\mathcal{W}}\phi{(x_{i}}))}]\\
&=\frac{1}{m}\sum\limits_{i = 1}^m [1 + \sum\limits_{u = 1}^K {\sum\limits_{v = 1}^K {\kappa ({w_u},{x_i})\kappa ({w_u},{w_v})\kappa ({w_v},{x_i})} }-\\
&\ \ \ 2\sum\limits_{u = 1}^K {{{(\kappa ({w_u},{x_i}))}^2}}  + \lambda \sum\limits_{j = 1}^K {\sqrt {\varepsilon  + \sum\limits_{u = 1}^K {{h_{ju}}{{(\kappa ({w_u},{x_i}))}^2}} } }],
\end{aligned}
\end{equation}
where $w_u$ and $w_v$ are the rows of basis $W$, and ${h_{ju}}$ is the element in pooling matrix $H$. Since the row $w_j$ of $W$ is contained in the kernel $\kappa ({w_j},\cdot)$, it is very hard to directly utilize the optimization methods in RICA, e.g. L-BFGS and CG~\cite{schmidt2005minfunc}, to compute the optimal basis. Thus, to solve this problem, we alternatively optimize each row of basis $W$ instead. With respect to each updating row $w_p$ of $W$, the derivative of $f(W)$ is

\begin{equation}
\label{equ:der}
\begin{aligned}
&\frac{{\partial f}}{{\partial{w_p}}} = \frac{{ - \gamma }}{m}\sum\limits_{i = 1}^m {[\sum\limits_{v = 1}^K {4\kappa ({w_p},{x_i})\kappa ({w_p},{w_v})\kappa ({w_v},{x_i})} }\\
& \times(({w_p} - {x_i})+ ({w_p} - {w_v})) - 8\kappa ({w_p},{x_i})({w_p} - {x_i}) \\
& + 2\lambda \sum\limits_{j = 1}^K {\frac{{{h_{jp}}\kappa ({w_p},{x_i})({w_p} - {x_i})}}{{\sqrt {\varepsilon  + \sum\limits_{v = 1}^K {{h_{jv}}{{(\kappa ({w_v},{x_i}))}^2}} } }}} ].
\end{aligned}
\end{equation}

Then, to compute the optimal $w_p$, we set $\frac{{\partial f}}{{\partial{w_p}}} =0$. Since $w_p$ is contained in $\kappa ({w_p},\cdot)$, it is challenging to solve the Equation (\ref{equ:der}). Thus, we seek the approximate solution instead of the exact solution. Inspired by fixed point algorithm~\cite{KSR_eccv10}, to update $w_p$ in the $(q)$-th iteration, we utilize the result of $w_p$ in the $(q - 1)$-th iteration to calculate the part in the kernel function. In addition, we utilize k-means to initialize the basis followed by ~\cite{KSR_eccv10}. Let denote the $w_p$ in the $(q)$-th iteration as $w_{p,(q)}$, and the Equation (\ref{equ:der}) with respect to $w_{p,(q)}$ becomes
\begin{align}
\label{equ:fp}
&\frac{{\partial f}}{{\partial w_{p,(q)}}}\cong  \frac{{ - \gamma }}{m}\sum\limits_{i = 1}^m {[\sum\limits_{v = 1}^K {4\kappa ({w_{p,{(q - 1)}}},{x_i})\kappa ({w_{p,{(q - 1)}}},{w_v})} }\times\nonumber\\
& \kappa ({w_v},{x_i})(({w_{p,(q)}} - {x_i}) + ({w_{p,(q)}} - {w_v}))- 8\kappa ({w_{p,(q - 1)}},{x_i}) \nonumber\\
& \times({w_{p,(q)}} - {x_i}) + 2\lambda \sum\limits_{j = 1}^K {\frac{{{h_{jp}}\kappa ({w_{p,(q - 1)}},{x_i})({w_{p,(q)}} - {x_i})}}{{\sqrt {\varepsilon  + \sum\limits_{v = 1}^K {{h_{jv}}{{(\kappa ({w_v},{x_i}))}^2}} } }}} ]\nonumber\\
&= 0. \nonumber
\end{align}
When all the remaining rows are fixed, the problem becomes a linear equation of $w_{p,(q)}$, which can be solved straightforwardly.
\subsection{Connection between kRICA and KSR}
It is clear there is a close connection between the proposed kRICA and KSR~\cite{KSR_eccv10}. Similar to kRICA, KSR attempts to find the sparse representation of nonlinear features in a high dimensional feature space and its optimization problem is
\begin{equation}
\label{equ:ksr}
\mathop {\min}\limits_{\mathcal{W},s_i} \frac{1}{m}\sum \limits_{i = 1}^m [||{{\mathcal{W}}^T}s_i - {\phi(x_{i})}||_2^2  + \lambda{||s_i||_1}],
\end{equation}
where $s_i\in R^K$ is the sparse representation of sample $x_i$. Therefore, there are two major differences between them.

(1) KSR utilizes explicit encoding for sparse representation corresponding to input data sample, i.e., $s_i={\mathcal{W}}\phi{(x_{i})}$. Since the objective of Equation (\ref{equ:ksr}) in KSR is not convex, the basis $\mathcal{W}$ and sparse codes $v_i$ should be optimized, alternatively.

(2) The simple $L_1$ penalty, $g(s_i)=||s_i||_1$, is employed by KSR to promote sparsity while kRICA uses $L_2$ pooling instead, which can force the pooling features to group similar features together to achieve invariance, and meanwhile optimize the sparsity.

\section{Supervised Kernel RICA}
Given the labeled training data, our goal is to utilize class information to learn a structured basis set, which is consisted of basis vectors from different basis subsets corresponding to different class labels. Then each subset will sparsely represent well for its own class but not for the others. Thus, to learn such basis, we further extend the unsupervised kRICA to a supervised one by introducing a discrimination constraint, namely d-kRICA.

Mathematically, when the sample $x_{i}$ is labeled as $y_{i} \in \{1,\ldots,c\}$ where $c$ is the total number of classes, we can further utilize class information to learn a structured basis set ${W}=[{W}^{(1)},{W}^{(2)},\ldots,{W}^{(c)}]^T$ $\in R^{K\times n}$, where ${W}^{({y_{i}})} \in R^{k \times n}$ is the basis subset that can well represent the sample $x_{i}$ belonging to the $y_{i}$-th class rather than others, $k$ is the number of basis vectors for each subset and $K=k*c$. Let denote $s_i = Wx_i$ where $s_i$ can be regarded as the sparse representation of sample $x_i$~\cite{Rica}.

\subsection{Discrimination constraint}
Since we aim to utilize class information to learn a structured basis, we hope that the sample $x_i$ labeled as $y_{i}$ will only be reconstructed by the basis subset $W_{y_i}$ with coefficients $s_i$. To achieve this goal, an inhomogeneous representation cost constraint ~\cite{dRICA,yang2011SRC} was utilized to minimize the inhomogeneous representation coefficients of $s_i$, i.e., coefficients corresponding to basis vectors other than belonging to $W_{y_i}$. However, this constraint only focuses on minimizing the inhomogeneous coefficients while fails to consider maximizing the the homogeneous ones, which is not sufficient to learn an optimal structured basis. Consequently, to learn such basis, we introduce a discrimination constraint, which maximizes the homogeneous representation cost and minimizes the inhomogeneous representation cost, jointly. Mathematically, we define the homogeneous cost as $P_+$ and the inhomogeneous cost as $P_-$. Specifically, $P_+$ and $P_-$ are
\begin{equation}
\label{equ:P}
\begin{aligned}
P_+ = || D_{+y_i}s_i ||_2^2,\\
P_- = || D_{-y_i}s_i ||_2^2,
\end{aligned}
\end{equation}
where $D_{+y_i}\in R^K$ and $D_{-y_i}\in R^K$ select the homogeneous and inhomogeneous representation coefficients of $s_i$, respectively.
For example, assuming $W=[W^{(1)},W^{(2)},W^{(3)}]^T$, $W^{(y_i)}\in R^{2 \times n}$($y_i\in\{1,2,3\}$) and ${y_i}$=3, $D_{+y_i}$ and $D_{-y_i}$ can be respectively defined as follows.
\begin{displaymath}
\begin{array}{l}
 {D_{+3}} = [\begin{array}{*{20}{c}}
   0 & 0 & 0 & 0 & 1 & 1  \\
\end{array}] \\
 {D_{-3}} = [\begin{array}{*{20}{c}}
   1 & 1 & 1 & 1 & 0 & 0  \\
\end{array}] \\
\end{array}
\end{displaymath}

Intuitively, we can define the discrimination constraint function $d(s_i)$ as $P_- - P_+$, which means the sparse representation $s_i$ in terms of basis matrix $W$ will only concentrate on the basis subset $W^{(y_i)}$.
However, this constraint is non-convex and unstable. To address the problem, we propose to incorporate an elastic term $||s_i||_2^2$ into $d(s_i)$. Thus, $d(s_i)$ is defined as
\begin{equation}
\label{equ:dx}
d(s_i) =|| D_{-y_i}s_i ||_2^2-|| D_{+y_i}s_i ||_2^2+\eta ||s_i||_2^2.
\end{equation}
It can be proved that if $\eta \geq k+1$, $d(s_i)$ is strictly convex to $s_i$. Please see the \textbf{Appendix B} for a detailed proof. The constraint (\ref{equ:dx}) maximizes the homogeneous representation cost and minimizes the inhomogeneous representation cost, simultaneously, which leads to learn a structured basis consisted of basis vectors from different basis subsets corresponding to the class labels. Then each subset will sparsely represent well for its own class but not for the others. Furthermore, data samples belonging to the same class will have similar representations, and thereby the obtained new representations can take more discriminative power.

By incorporating the discrimination constraint into the kRICA framework (d-kRICA), we can get the following objective function
\begin{equation}
\label{equ:d-krica}
\begin{aligned}
\mathop {\min}\limits_\mathcal{W} &\frac{1}{m}\sum \limits_{i = 1}^m [{||{{\mathcal{W}}^T}{\mathcal{W}}{\phi(x_{i}}) - {\phi(x_{i})}||_2^2 } + \\
&\lambda{g({\mathcal{W}}\phi{(x_{i}}))}+ \alpha d({\mathcal{W}}\phi{(x_{i}}))],
\end{aligned}
\end{equation}
where $\lambda$ and $\alpha$ are the scalars controlling the relative contribution of the corresponding terms. Given a test sample, Equation (\ref{equ:d-krica}) means that the learned basis set can sparsely represent it with nonlinear structure while demands its homogeneous representations as large as possible and meanwhile inhomogeneous representations as small as possible. Following kRICA, the optimization problem (\ref{equ:d-krica}) can be easily solved by the above proposed fixed point algorithm.

\section{Experiments}
In this section, we will firstly introduce the feature extraction for image classification. Then, we evaluate the performances of our kRICA and d-kRICA for image classification on three public datasets: Caltech 101~\cite{Caltech101}, CIFAR-10~\cite{coates2010analysis} and STL-10~\cite{coates2010analysis}. Furthermore, we study the selections of tuning parameters  and kernel functions for our method. Finally, we give the similarity matrix to further illustrate the performances of kRICA and d-kRICA.
\subsection{Feature Extraction for Classification}

Given a $p\times p$ input image patch (with $d$ channels) $x \in R^n$ ($n=p \times p \times d$), kRICA can transform it to a new representation $s={\mathcal{W}}{\phi(x_{i})} \in R^K$ in the feature space, where $p$ is termed as the 'receptive field size'. For an image of $N\times M$ pixels (with $d$ channels), we could obtain a $(N - p + 1) \times (M - p + 1)$(with $K$ channels) feature following the same setting in~\cite{Rica}, by estimating the representation for each $p\times p$ 'subpatch' of the input image. To reduce the dimensionality of the image representation, we utilize similar pooling method in~\cite{Rica} to form a reduced $4K$-dimensional pooled representation for image classification. Given the pooled feature for each image, we utilize linear SVM for classification.

\subsection{Classification on Caltech 101}
\label{sec:Caltech101}
Caltech 101 dataset consists of 9144 images which are divided among 101 object classes and 1 background class including animals, vehicles, etc. Following the common experiment setup~\cite{scspm}, we implement our algorithm on 15 and 30 training images per category with basis size $K = 1020$ and 10$\times$10 receptive fields, respectively. Comparison results are shown in Table 2. We compare our classification accuracy with ScSPM~\cite{scspm}, D-KSVD~\cite{DKSVD}, LC-KSVD~\cite{lcksvd}, RICA~\cite{Rica}, KICA~\cite{kica}, KSR~\cite{KSR_eccv10} and DRICA~\cite{dRICA}. In addition, in order to compare with DRICA, we incorporate the discrimination constraint (\ref{equ:dx}) into the RICA framework (\ref{rica}), namely d-RICA.

Table~\ref{tab:cal} shows that kRICA and d-kRICA outperform the other competing approaches.

\begin{table}[tb]
\caption{Image classification Accuracy on Caltech 101 dataset.}
\label{tab:cal}
\begin{center}
\begin{tabular}{l|c|c}\hline

Training size                   & 15          & 30  \\
\hline\hline
ScSPM~\cite{scspm}              & 67.0\%      & 73.2\%\\
D-KSVD~\cite{DKSVD}             & 65.1\%      & 73.0\%\\
LC-KSVD~\cite{lcksvd}           & 67.7\%      & 73.6\% \\
RICA~\cite{Rica}                & 67.1\%      & 73.7\% \\
KICA~\cite{kica}                & 65.2\%      &72.8\% \\
KSR~\cite{KSR_eccv10}           & 67.9\%      & 75.1\% \\
DRICA~\cite{dRICA}              & 67.8\%      & 74.4\% \\
\hline
d-RICA                          & 68.7\%      & 75.6\% \\
kRICA                           & 68.2\%      & 75.4\% \\
d-kRICA                         &\textbf{71.3\%}       &\textbf{77.1\%} \\
\hline
\end{tabular}
\end{center}
\vspace{-3ex}
\end{table}%

\subsection{Classification on CIFAR-10}

The CIFAR-10 dataset includes 10 categories and 60000 32$\times$32 color images in all with 6000 images per category, such as airplane, automobile, truck and horse etc. In addition, there are 50000 training images and 10000 testing images. Specifically, 1000 images from each class are randomly selected as test images and the other 5000 images from each class as training images. In this experiment, we fix the size of basis set to 4000 with 6$\times$6 receptive fields followed by~\cite{coates2010analysis}. We compare our approach with RICA, K-means (Triangle, 4000 features)~\cite{coates2010analysis}, KSR, DRICA and d-RICA etc.

Table~\ref{tab:cifar} shows the effectiveness of our proposed kRICA and d-kRICA.
\begin{table}
\centering
\caption{Test Classification Accuracy on CIFAR-10 dataset.}
\begin{tabular}{l|c} \hline
Model & Accuracy\\ \hline \hline
Improved Local Coord. Coding~\cite{ILCC} &74.5\% \\
Conv. Deep Belief Net (2 layers)~\cite{krizhevsky2010convolutional} &78.9\%\\
Sparse auto-encoder~\cite{coates2010analysis} &73.4\%\\
Sparse RBM~\cite{coates2010analysis} &72.4\%\\
K-means (Hard)~\cite{coates2010analysis} &68.6\%\\
K-means (Triangle)~\cite{coates2010analysis} &77.9\%\\
K-means (Triangle, 4000 features)~\cite{coates2010analysis} &79.6\%\\
RICA~\cite{Rica}             &81.4\%\\
KICA~\cite{kica}             &78.3\% \\
KSR~\cite{KSR_eccv10}        &82.6\% \\
DRICA~\cite{dRICA}           &82.1\%\\
\hline
d-RICA                       &82.9\% \\
kRICA                        &83.4\% \\
d-kRICA                      &\textbf{84.5\%} \\
\hline
\end{tabular}
\label{tab:cifar}
\end{table}

\subsection{Classification on STL-10}
In STL-10, there are 10 classes(e.g., airplane, dog, monkey and ship etc), where each image is 96x96 pixels and color. In addition, this dataset is divided into 500 training images (10 pre-defined folds), 800 test images per class and 100,000 unlabeled images for unsupervised learning. In our experiments, we set the size of basis set $K$= 1600 and 8$\times$8 receptive fields in the same manner described in~\cite{Rica}.

Table~\ref{tab:stl} shows the classification results of the raw pixels~\cite{coates2010analysis}, K-means, RICA, KSR, DRICA, d-RICA, kRICA and d-kRICA.

As can be seen, d-RICA achieves better performance than DRICA on all of the above datasets.
It is because that DRICA just only minimized the inhomogeneous representation cost for structured basis learning, while d-RICA simultaneously maximizes the homogeneous representation cost and minimizes the inhomogeneous representation cost, which makes the learned sparse representation take more discriminative power.
Although both DRICA and d-RICA introduce the class information, unsupervised kRICA still performs better than both these algorithms. This means that kRICA implies more discriminative power for classification by representing the data with nonlinear structure. Additionally, since kRICA utilizes the $L_2$ pooling instead of $L_1$ penalty to achieve feature invariance, it demonstrates better performance than KSR. Furthermore, the d-kRICA achieves better performance than kRICA in all the cases by bringing in class information.

\begin{figure}
\centering
\psfig{file=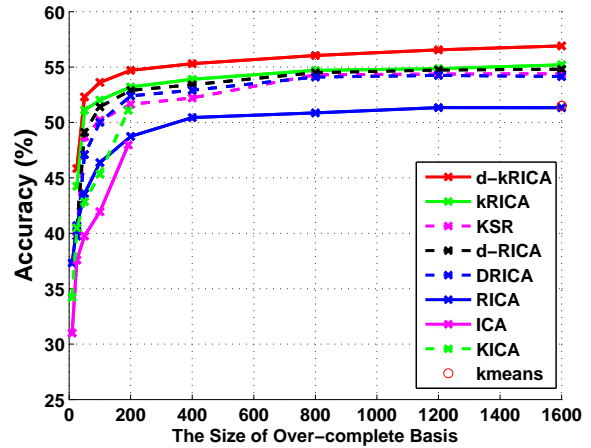, height=2.5in}
\caption{Classification performance on STL-10 dataset with varying basis size and 8$\times$8 receptive fields.}
%\vskip -6pt
\label{fig:dic}
\end{figure}

We also investigate the effect of basis size for our proposed kRICA and d-kRICA on STL-10 dataset. In our experiments, we try seven sizes: 50, 100, 200, 400, 800, 1200 and 1600. As shown in Fig.~\ref{fig:dic}, the classification accuracies of d-kRICA and kRICA continue to increase when the basis size goes up to 1600 and the performances augment slightly from basis size of 800. Especially, d-kRICA outperforms all the other algorithms all the way.

\begin{table}
\centering
\caption{Test Classification Accuracy on STL-10 dataset.}
\begin{tabular}{l|c} \hline
Model & Accuracy\\ \hline \hline
Raw pixels~\cite{coates2010analysis} &31.8\%\\
K-means(Triangle 1600 features)~\cite{coates2010analysis}    &51.5\%\\
RICA(8x8 receptive fields)~\cite{Rica}   &51.4\%\\
RICA(10x10 receptive fields)~\cite{Rica}   &52.9\%\\
KICA~\cite{kica}             &51.1\% \\
KSR~\cite{KSR_eccv10}        &54.4\% \\
DRICA~\cite{dRICA}           &54.2\%\\
\hline
d-RICA          &54.8\% \\
kRICA           &55.2\% \\
d-kRICA         &\textbf{56.9\%} \\
\hline
\end{tabular}
\label{tab:stl}
\end{table}
\subsection{Tuning Parameter and Kernel Selection}
In the experiments, the tuning parameters in kRICA and d-kRICA, i.e. $\lambda$, $\alpha$ and $\gamma$ in the objective function, are verified by cross validation to avoid over-fitting. More specifically, we experimentally set these parameters as follows.

\begin{figure}
\centering
\psfig{file=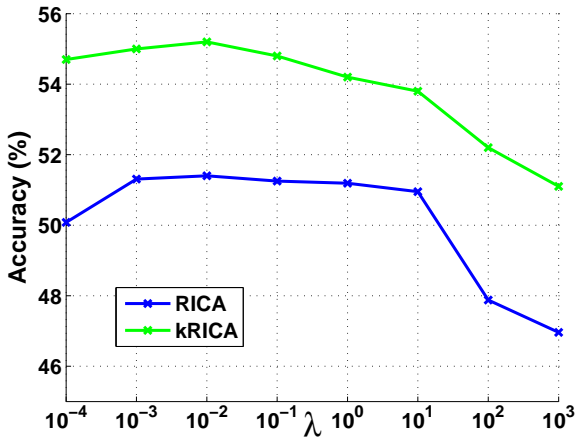, height=2.5in}
\caption{The relationship between the weight of sparsity term ($\lambda$) and classification accuracy on STL-10 dataset.}
%\vskip -6pt
\label{fig:lambda}
\end{figure}

\textbf{The effect of $\boldsymbol{\lambda}$} : The parameter $\lambda$ is the weight of sparsity term, which is an important factor in kRICA. To facilitate the parameter selection, we experimentally investigate how the performance of kRICA varies with the parameter $\lambda$ on STL-10 dataset in Fig.~\ref{fig:lambda} ($\gamma = 10^{-1}$).
Fig.~\ref{fig:lambda} shows that kRICA achieves best performance when $\lambda$ is fixed to be $10^{-2}$. Thus, we set $\lambda=10^{-2}$ for STL-10 data. In addition, we test the accuracy of RICA under the same sparsity weight. It is easy to find that our proposed nonlinear RICA (kRICA) can consistently outperform linear RICA with respect to $\lambda$. Similarly, we experimentally set $\lambda=10^{-1}$ for Caltech data and $\lambda=10^{-2}$ for CIFAR-10 data.

\begin{figure}
\centering
\psfig{file=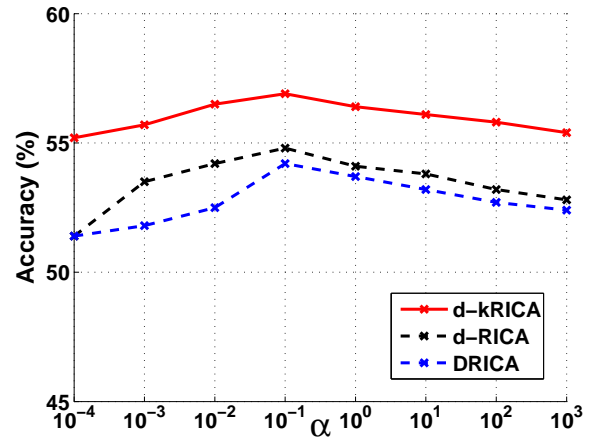, height=2.5in}
\caption{The relationship between the weight of discrimination constraint term ($\alpha$) and classification accuracy on STL-10 dataset.}
%\vskip -6pt
\label{fig:alpha}
\end{figure}

\textbf{The effect of} $\boldsymbol{\alpha}$ : The parameter $\alpha$ controls the weight of discrimination constraint term. When $\alpha=0$, the supervised d-kRICA optimization problem becomes the unsupervised kRICA problem. Fig.~\ref{fig:alpha} shows the relationship between the weight of discrimination constraint term $\alpha$ and classification accuracy on the STL-10. We can see that d-kRICA achieves best performance when $\alpha=10^{-1}$. Hence, we set $\alpha=10^{-1}$ for STL-10 data.
In particular, d-RICA achieves better performance than DRICA in a wide range of $\alpha$ values. This is because that DRICA just only minimizes the inhomogeneous representation cost, while d-RICA jointly optimizes both the homogeneous and inhomogeneous representation costs for basis learning, which makes the learned sparse representations take more discriminative power.
Furthermore, by representing the data with nonlinear structure, d-kRICA implies more discriminative power for classification and outperforms both these algorithms. Similarly, we set $\alpha=1$ for Caltech data and $\alpha=10^{-1}$ for CIFAR-10 data.

\begin{figure}
\centering
\psfig{file=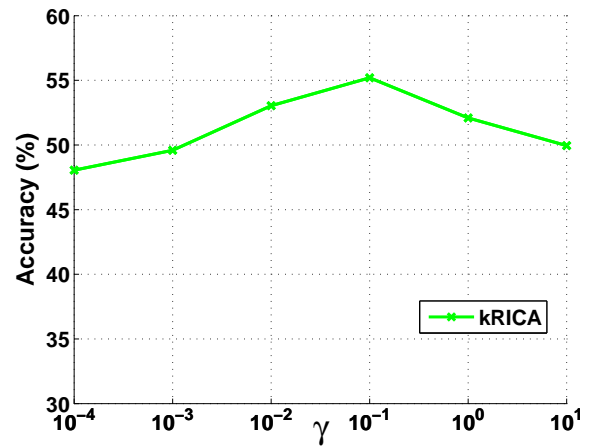, height=2.5in}
\caption{Classification performance on STL-10 dataset with varying kernel parameter ($\gamma$) in Gaussian kernel.}
%\vskip -6pt
\label{fig:gamma}
\end{figure}

\textbf{The effect of $\boldsymbol{\gamma}$} :  When we utilize the Gaussian kernel in kRICA, it is vital to select the kernel parameter $\gamma$, which affects the image classification accuracy. Fig.~\ref{fig:gamma} shows the relationship between $\gamma$ and classification accuracy on STL-10 dataset. Therefore, we set $\gamma=10^{-1}$ for STL-10 data. Similarly, we experimentally set $\gamma=10^{-2}$ for Caltech data and $\gamma=10^{-1}$ for CIFAR-10 data.

We also investigate the effect of different kernels for kRICA in image classification, i.e., Polynomial kernel: $(1+x^Ty)^b$, Inverse Distance kernel: $\frac{1}{1+b||x-y||}$, Inverse Square Distance kernel: $\frac{1}{1+b||x-y||^2}$, Exponential Histogram Intersection kernel: $\sum_i\min(e^{bx_i},e^{by_i})$.\footnote{Following the work~\cite{KSR}, we set b=3 for Polynomial kernel and b=1 for the others.} Table~\ref{tab:kernel} demonstrates the classification performances of different kernels on STL-10 dataset, and Gaussian kernel outperforms the other kernels. Thus, we employ Gaussian kernel in our studies.

\begin{table}
\centering
\caption{Classification performances of different kernels on STL-10 dataset.}
\begin{tabular}{l|c} \hline
Kernel & Accuracy\\ \hline \hline
Polynomial kernel  &54.2\%\\
Inverse Distance kernel    &38.3\%\\
Inverse Square Distance kernel   &47.6\%\\
Exponential Histogram Intersection kernel  &36.5\%\\
Gaussian kernel         &\textbf{56.9\%} \\
\hline
\end{tabular}
\label{tab:kernel}
\end{table}
\subsection{Similarity Analysis}
\begin{figure*}
  \centering
  \subfigure[RICA]
  {
    \label{fig:subfig:a} %% label for first subfigure
    \includegraphics[width=2.3 in]{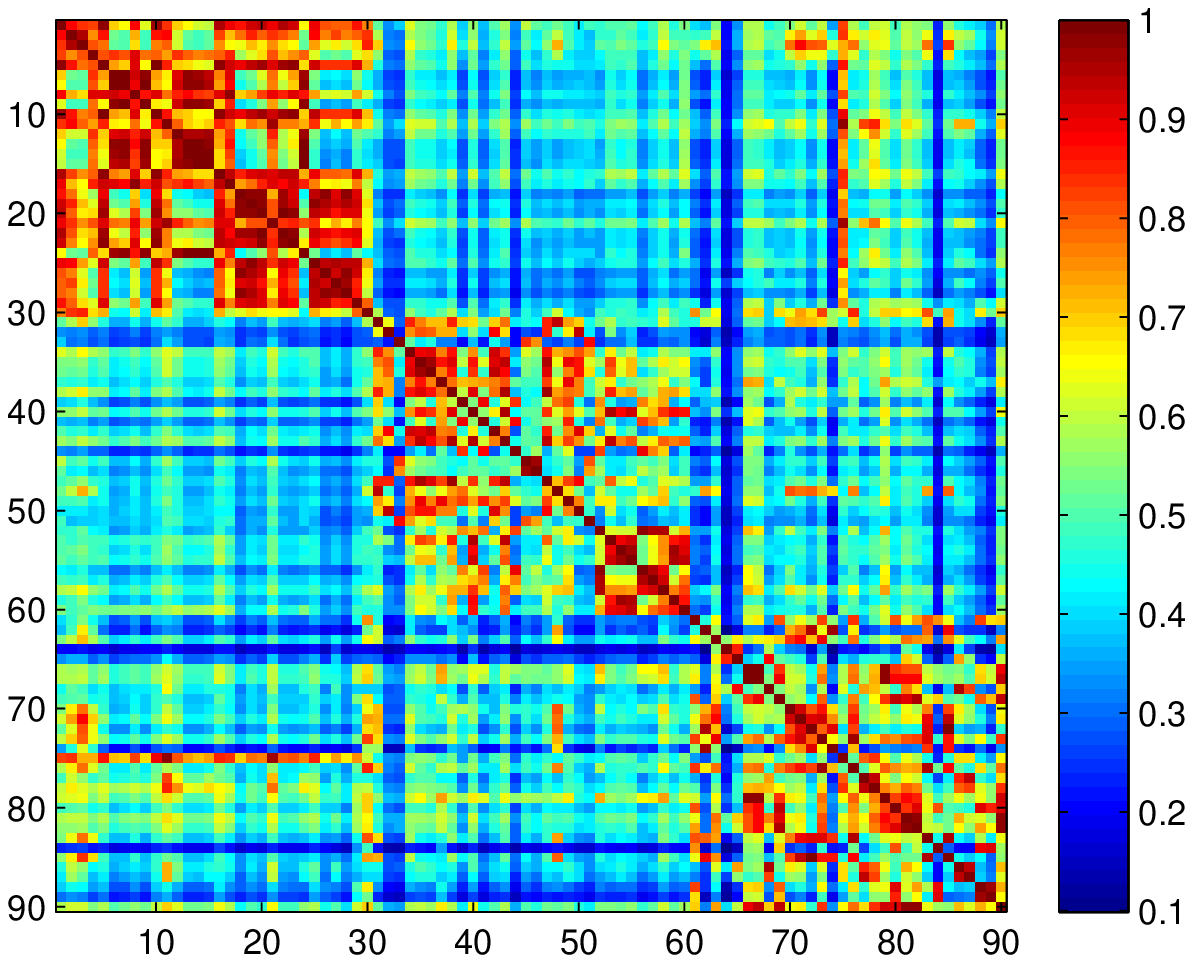}
  }
  \hspace{-0.35in}
  \subfigure[kRICA]
  {
    \label{fig:subfig:b} %% label for second subfigure
    \includegraphics[width=2.3 in]{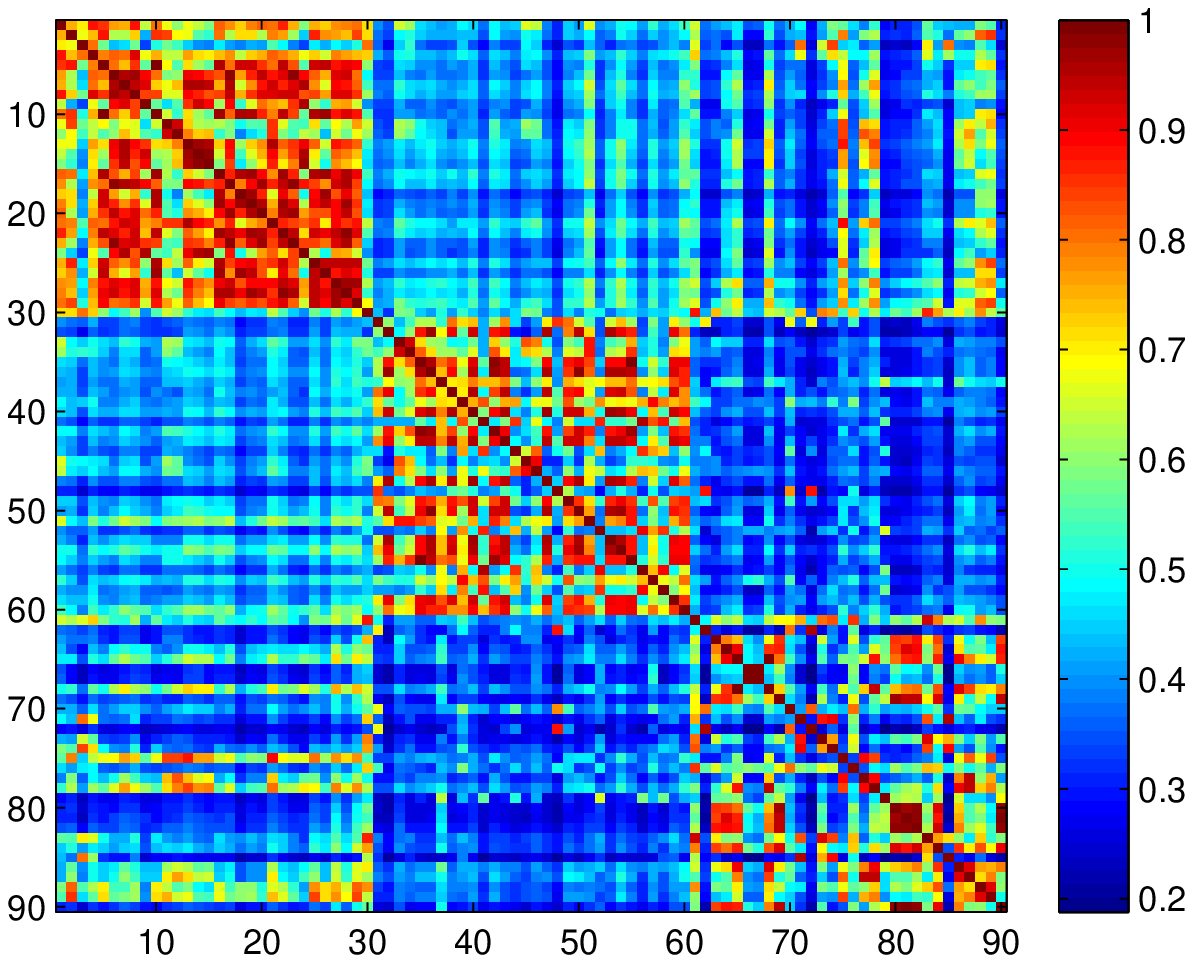}
  }
  \hspace{-0.35in}
  \subfigure[d-kRICA]
  {
    \label{fig:subfig:b} %% label for second subfigure
    \includegraphics[width=2.3 in]{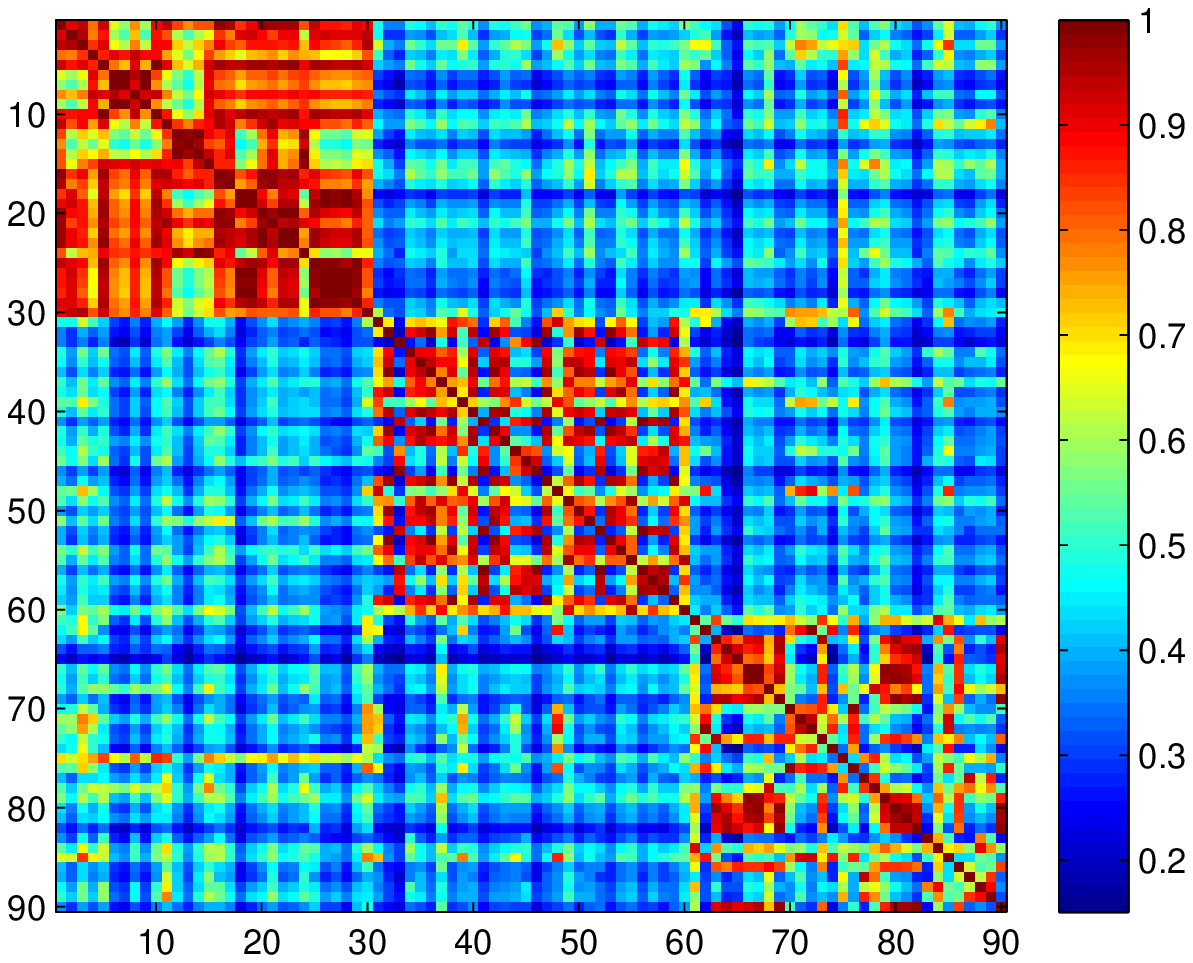}
  }
  \caption{The similarity matrices for sparse representations of RICA, kRICA and d-kRICA.}
  \label{fig:simi} %% label for entire figure
\end{figure*}

In above sections, we have shown the effectiveness of kRICA and d-kRICA for image classification. To further illustrate their performances, we firstly choose 90 images from three classes in Caltech 101, and 30 images for each class. Then we compute the similarity between sparse representations of these images for RICA, kRICA and d-kRICA, respectively. Fig.~\ref{fig:simi} demonstrates the similarity matrices corresponding to sparse representations of RICA, kRICA and d-kRICA, respectively. Each element $(i, j)$ in similarity matrix is the sparse representation similarity measured by Euclidean distance between image $i$ and $j$.
Since a good sparse representation method can make the new representations belonging to the same class more similar, their similarity matrix also should be block-wise. Fig.~\ref{fig:simi} shows that nonlinear kRICA takes more discriminative power than linear RICA, and d-kRICA achieves best by binging in class information.

\section{Conclusions}

In this paper, we propose a kernel ICA model with reconstruction constraint (kRICA) to capture the nonlinear features. To bring in the class information, we further extend the unsupervised kRICA to a supervised one by introducing a discrimination constraint. This constraint leads to learn a structured basis consisted of basis vectors from different basis subsets corresponding to different class labels. Then each subset will sparsely represent well for its own class but not for the others. Furthermore, data samples belonging to the same class will have similar representations, and thereby the obtained sparse representation can take more discriminative power. The experiments conducted on standardized datasets have demonstrated the effectiveness of our proposed method.

\appendices
\section{Proof of Lemma 4.1}
\textbf{Poof}

Since the input data set $X=\{ {x_{i}}\} _{i = 1}^m$ is whitened in the feature space by KPCA, we have
\begin{align}
{E}[\phi(X)\phi(X)^T]=\frac{1}{m}\sum \limits_{i = 1}^m \phi(x_i)\phi(x_i)^T=I,\nonumber
\end{align}
where $I$ is the identity matrix. Furthermore, we can obtain
\begin{align}
&\frac{1}{m}\sum \limits_{i = 1}^m {||{{\mathcal{W}}^T}{\mathcal{W}}{\phi(x_{i}}) - {\phi(x_{i})}||_2^2 }\nonumber\\
=&\frac{1}{m}\sum \limits_{i = 1}^m Tr[({{\mathcal{W}}^T}{\mathcal{W}}{\phi(x_{i}}) - {\phi(x_{i})})^T({{\mathcal{W}}^T}{\mathcal{W}}{\phi(x_{i}}) - {\phi(x_{i})})]\nonumber\\
=&Tr[({{\mathcal{W}}^T}{\mathcal{W}} - I)^T({{\mathcal{W}}^T}{\mathcal{W}} -I)\frac{1}{m}\sum \limits_{i = 1}^m\phi(x_i)\phi(x_i)^T]\nonumber\\
=&{||{{\mathcal{W}}^T}{\mathcal{W}} - I||_\mathcal{F}^2 },\nonumber
\end{align}
where ${Tr}[\cdot]$ denotes the trace of a matrix, and the steps of derivation employ the matrix property ${Tr}(AB)={Tr}(BA)$. Thus, the reconstruction cost is equivalent to the orthonormality constraint when data is whitened in the feature space.
\section{Proof of the convexity of $d(s_i)$}
We rewrite the Equation (\ref{equ:dx}) as
\begin{equation}
\begin{aligned}
d(s_i)&=|| D_{-y_i}s_i ||_2^2-|| D_{+y_i}s_i ||_2^2+\eta ||s_i||_2^2\\
&=Tr[s_i^T D_{{-y_i}}^T D_{{-y_i}}s_i-s_i^TD_{{+y_i}}^T D_{{+y_i}}s_i+ \eta s_i^T s_i]
\end{aligned}
\end{equation}
Then, we can obtain its Hessian matrix $\nabla^2 d$ with respect to $ s_i$.
\begin{equation}
\begin{aligned}
\nabla^2 d =& 2D_{{y_i-}}^T D_{{y_i-}}-2D_{{y_i+}}^T D_{{y_i+}}+2\eta I
\end{aligned}
\end{equation}
Without loss of generality, we assume
\begin{align}
D_{+y_i} = [{0\ \cdots \ 0\  {\mathop {\overbrace{1\ \cdots \ 1}} \limits^k}\ 0\ \cdots \ 0\ }] \in R^{K}, \nonumber\\
D_{-y_i} = [{1\ \cdots \ 1\  {\mathop {\overbrace{0\ \cdots \ 0}} \limits^k}\ 1\ \cdots \ 1\ }] \in R^{K}. \nonumber
\end{align}
After some derivations, we have
$\nabla^2 d = 2\times A,$
where
{\footnotesize{
\begin{displaymath}
\setlength{\arraycolsep}{1.5pt}
A=\left[ {\begin{array}{*{9}{c}}
   {\eta  + 1}   &  \cdots  & 1 & 0 &  \cdots  & 0 & 1 &  \cdots  & 1  \\
    \vdots  &  \vdots  &  \vdots  &  \vdots  &  \vdots  &  \vdots  &  \vdots  &  \vdots  &  \vdots   \\
   1 &  \cdots  & {\eta  + 1} & 0 &  \cdots  & 0 & 1 &  \cdots  & 1  \\
   0 &  \cdots  & 0 & {\eta  - 1} &  \cdots  & { - 1} & 0 &  \cdots  & 0  \\
    \vdots  &  \vdots  &  \vdots  &  \vdots  &  \vdots  &  \vdots  &  \vdots  &  \vdots  &  \vdots   \\
   0 &  \cdots  & 0 & { - 1} &  \cdots  & {\eta  - 1} & 0 &  \cdots  & 0  \\
   1 &  \cdots  & 1 & 0 &  \cdots  & 0 & {\eta  + 1} &  \cdots  & 1  \\
    \vdots  &  \vdots  &  \vdots  &  \vdots  &  \vdots  &  \vdots  &  \vdots  &  \vdots  &  \vdots   \\
   1 &  \cdots  & 1 & 0 &  \cdots  & 0 & 1 &  \cdots  & {\eta  + 1}  \\
\end{array}} \right].
\end{displaymath}}}

The convexity of $d(s_i)$ depends on whether its Hessian matrix $\nabla^2 d$, i.e. matrix $A$, is positive definite or not~\cite{convex}. Meanwhile, the $K \times K$ matrix $A$ is positive definite if and only if $z^T A z>0$ for all nonzero vectors $z \in R^K$~\cite{matrix}, where $z^T$ denotes the transpose. Let the size of upper left matrix in $A$ be $t\times t$, and suppose $z=[z_1,\cdots,z_t,z_{t+1},\cdots,z_{t+k},z_{t+k+1},\cdots,z_{K}]^T$.
Then, we have {\footnotesize{
\begin{displaymath}
Az = \left[ {\begin{array}{*{20}{c}}
   {(\eta  + 1){z_1} + {z_2} +  \cdots  + {z_t} + {z_{t + k + 1}} +  \cdots  + {z_K}}  \\
    \vdots   \\
   {{z_1} + {z_2} +  \cdots  + (\eta  + 1){z_t} + {z_{t + k + 1}} +  \cdots  + {z_K}}  \\
   {(\eta  - 1){z_{t + 1}} - {z_{t + 2}} -  \cdots  - {z_{t + k}}}  \\
    \vdots   \\
   { - {z_{t + 1}} - {z_{t + 2}} -  \cdots  + (\eta  - 1){z_{t + k}}}  \\
   {{z_1} + {z_2} +  \cdots  + {z_t} + (\eta  + 1){z_{t + k + 1}} +  \cdots  + {z_K}}  \\
    \vdots   \\
   {{z_1} + {z_2} +  \cdots  + {z_t} + {z_{t + k + 1}} +  \cdots  + (\eta  + 1){z_K}}  \\
\end{array}} \right].
\end{displaymath}}}

Furthermore, we can get

{\footnotesize{
\begin{displaymath}
\setlength{\arraycolsep}{1pt}
\begin{aligned}
 {z^T}Az = &
 (\eta  + 1)\sum\limits_{i = 1}^t {{\rm{z}}_i^2}  + (\eta  - 1)\!\!\sum\limits_{i = t + 1}^{t + k}\!\!{{\rm{z}}_i^2}+ (\eta  + 1)\!\!\!\!\sum\limits_{i = t + k + 1}^K \!\!\!\!\!{{\rm{z}}_i^2}  \\
   &+ \quad 2\!\!\!\!\sum\limits_{\scriptstyle 1 \le i \le t - 1 \hfill \atop
  {\scriptstyle 2 \le j \le t \hfill \atop
  \scriptstyle i < j \hfill}}\!\!\!\!\!\!\! {{z_i}{z_j}} \ \ + \quad 2\!\!\!\!\!\!\!\!\sum\limits_{\scriptstyle 1 \le i \le t \hfill \atop \scriptstyle t + k + 1 \le j \le K \hfill}\!\!\!\!\!\!\!\!\!\!\! {{z_i}{z_j}} \ \ + \ \quad 2\!\!\!\!\!\!\!\!\!\!\!\!\sum\limits_{\scriptstyle t + 1 \le i \le t + k - 1 \hfill \atop
  {\scriptstyle t + 2 \le j \le t + k \hfill \atop   \scriptstyle i < j \hfill}}\!\!\!\!\!\!\!\!\!\!\!\!\!{{z_i}{z_j}}  \\
  &-\quad 2\!\!\!\!\!\!\!\!\!\!\!\sum\limits_{\scriptstyle t + 1 \le i \le t + k - 1 \hfill \atop
  {\scriptstyle t + 2 \le j \le t + k \hfill \atop \scriptstyle i < j \hfill}}\!\!\!\!\!\!\!\!\!\!\!\!\!{{z_i}{z_j}}  \\
  = &\eta (\sum\limits_{i = 1}^t {{\rm{z}}_i^2}  + \sum\limits_{i = t + k + 1}^K \!\!\!\!\!\!{{\rm{z}}_i^2} ) + ({z_1} + {z_2} +  \cdots  + {z_t} + {z_{t + k + 1}} \\
  &+  \cdots  + {z_K})^2  + (\eta  - 1)\!\!\sum\limits_{i = t + 1}^{t + k} {{\rm{z}}_i^2} \ \  -\quad 2\!\!\!\!\!\!\!\!\!\!\sum\limits_{\scriptstyle t + 1 \le i \le t + k - 1 \hfill \atop {\scriptstyle t + 2 \le j \le t + k \hfill \atop \scriptstyle i < j \hfill}}\!\!\!\!\!\!\!\!\!\!\!\! {{z_i}{z_j}}.
 \end{aligned}
 \end{displaymath}}}
 Define function $h(\eta)={z^T}Az$, and when $\eta \geq k+1$, it is easy to verify that
 {\footnotesize{
 \begin{displaymath}
 \begin{aligned}
& h(\eta ) \ge h(k+1) = (k+1)(\sum\limits_{i = 1}^t {{\rm{z}}_i^2} \  + \sum\limits_{i = t + k + 1}^K \!\!\!\!\!{{\rm{z}}_i^2} )+({z_1} +   \cdots  + {z_t}  \\
&+ {z_{t + k + 1}}+  \cdots  + {z_K})^2   + k\sum\limits_{i = t + 1}^{t + k} {{\rm{z}}_i^2}
\ -\quad 2\!\!\!\!\!\!\!\!\!\!\sum\limits_{\scriptstyle t + 1 \le i \le t + k - 1 \hfill \atop
  {\scriptstyle t + 2 \le j \le t + k \hfill \atop \scriptstyle i < j \hfill}}\!\!\!\!\!\!\!\!\!\!\!\!\!{{z_i}{z_j}}  \\
&= k(\sum\limits_{i = 1}^t {{\rm{z}}_i^2}\  +  \sum\limits_{i = t + k + 1}^K\!\!\!\!\!\!{{\rm{z}}_i^2} )+({z_1} +   \cdots  + {z_t} + {z_{t + k + 1}} +  \cdots   + {z_K})^2 \\
&+ \sum\limits_{i = 1}^{K}{{\rm{z}}_i^2} \ \  +\  {\sum\limits_{\scriptstyle t + 1 \le i \le t + k - 1 \hfill \atop
  {\scriptstyle t + 2 \le j \le t + k \hfill \atop  \scriptstyle i < j \hfill}}\!\!\!\!\!\!\!\!\!\!\!\!\!\!{({z_i} - {z_j})} ^2}.
 \end{aligned}
  \end{displaymath}}}
  Since $\sum\limits_{i = 1}^{K} {{\rm{z}}_i^2}>0$, we have $ h(\eta ) \ge h(k+1)>0$. Thus, Hessian matrix $\nabla^2 d$ is positive definite for $\eta \geq k+1$, which guarantees that $d(s_i)$ is convex to $s_i$.
\ifCLASSOPTIONcompsoc
  \section*{Acknowledgments}
\else
  \section*{Acknowledgment}
\fi

We thank Wende Dong for helpful discussions, and acknowledge Quoc V. Le for providing the RICA code.

\ifCLASSOPTIONcaptionsoff
  \newpage
\fi

\bibliographystyle{IEEEtran}
\bibliography{refs}

\end{document}